\documentclass[9pt,shortpaper,twoside,web]{ieeecolor}
\usepackage{generic}
\usepackage{cite}
\usepackage{amsmath,amssymb,amsfonts}
\usepackage{algorithmic}
\usepackage{graphicx}
\usepackage{textcomp}
\usepackage{color}
\usepackage{url}
\usepackage{booktabs}

\newcommand*{\tc}[1]{{\color{black}#1}}
\newcommand*{\forceblack}[1]{{\color{black}#1}}

\def\BibTeX{{\rm B\kern-.05em{\sc i\kern-.025em b}\kern-.08em
    T\kern-.1667em\lower.7ex\hbox{E}\kern-.125emX}}
\begin{document}
\title{Real-Time Gait Phase and Task Estimation for Controlling\\ a Powered Ankle Exoskeleton on Extremely Uneven Terrain}
\author{Roberto Leo Medrano, \IEEEmembership{Student Member, IEEE}, Gray Cortright Thomas, \IEEEmembership{Member, IEEE},\\ Connor G. Keais, Elliott J. Rouse, \IEEEmembership{Senior Member, IEEE} and Robert D. Gregg, \IEEEmembership{Senior Member, IEEE}
\thanks{This work was supported by the National Institute of Child Health \& Human Development of the NIH under Award Number R01HD094772 and by the D. Dan and Betty Kahn Foundation. The content is solely the responsibility of the authors and does not necessarily represent the official views of the NIH. \textit{Corresponding author: Robert Gregg}}
\thanks{R. Medrano and E. Rouse are with the Department of Mechanical Engineering, and G. Thomas, C. Keais, E. Rouse, and R. Gregg are with the Department of Robotics, University of Michigan, Ann Arbor, MI 48109. Contact: {\tt\small \{leomed,gcthomas,ckeais,ejrouse,rdgregg\}@umich.edu}}%
}

\maketitle

\begin{abstract}
Positive biomechanical outcomes have been reported with lower-limb exoskeletons in laboratory settings, but these devices have difficulty delivering appropriate assistance in synchrony with human gait as the task or rate of phase progression change in real-world environments. \tc{This paper presents a controller for an ankle exoskeleton that uses a data-driven kinematic model to continuously estimate the phase, phase rate, stride length, and ground incline states during locomotion, which enables the real-time adaptation of torque assistance to match human torques observed in a multi-activity database of 10 able-bodied subjects.} \tc{We demonstrate in live experiments with a new cohort of 10 able-bodied participants that the controller yields phase estimates comparable to the state of the art, while also estimating task variables with similar accuracy to recent machine learning approaches. The implemented controller successfully adapts its assistance in response to changing phase and task variables, both during controlled treadmill trials (}\tc{N=10, phase RMSE: 4.8 $\pm$ 2.4\%}) and a real-world stress test with extremely uneven terrain \tc{(N=1, phase RMSE: 4.8 $\pm$ 2.7\%).}
\end{abstract}


\section{Introduction}
\label{sec:introduction}

Robotic exoskeletons may someday allow us to overcome the limits of our natural bodies. Emerging lower-limb exoskeletons are capable of providing assistive joint torques to help their wearers walk and carry loads with promising outcomes, including reduced metabolic cost \cite{Mooney2014b,Zhang2017,Ding2018,Panizzolo2019,Sawicki2020,Mooney2014} and muscular effort \cite{gordon2007learning,lenzi2013powered,Zhu2021}. Most research to date has focused on steady-state locomotion in a controlled laboratory setting where the task and phase rate (rate of continuous progression through the gait cycle) are nearly constant. This regulated environment makes it easier to design control strategies that deliver appropriate torque assistance in synchrony with the user's gait. However, control strategies based on these assumptions perform poorly outside of the laboratory, where environments are uncertain and locomotion is highly non-steady and transitory. In order for the field to study biomechanical outcomes outside of the laboratory, new control strategies are needed that explicitly account for continuously varying task and phase.

During steady-state locomotion in controlled laboratory settings, phase progression can be reasonably predicted using time normalized by the stride period. The stride period is usually estimated as the time between subsequent ipsilateral heel strike (HS) events, which can reasonably predict the next HS event during steady locomotion. This `timing-based' approach is quite effective and widely used for controlling exoskeletons on treadmills \cite{Tucker2015}. Typically this phase estimate parameterizes a pre-defined torque profile to deliver real-time assistance through the exoskeleton's actuator(s). Recent work has demonstrated impressive reductions in the metabolic rate of level-ground treadmill walking by optimizing this torque profile in real-time \cite{Zhang2017,Ding2018}. While this paradigm of timing-based estimation (TBE) for torque control works well in steady-state locomotion, it is not designed for more practical conditions outside the laboratory where both the periodicity of gait and the task can change sporadically.

Recent work has addressed the problem of \emph{non-constant} phase progression in powered prostheses and orthoses by introducing phase-based\footnote{We define `phase-based' approaches as those where the rate of phase progression can change continuously within a stride, whereas `timing-based' approaches have a fixed rate over the stride. Note that both would fall under the category of `phase-based' approaches according to \cite{Tucker2015}.} controllers \cite{Holgate2009,Quintero2018a, Rezazadeh2019, Kang2019, Kang2021, Seo2019,hong2021phase,zhang2021adaptive,shepherd2022deep}, which continuously adjust the rate of phase progression to accommodate dynamic speed changes mid-stride. In contrast to the timing-based approach, these controllers estimate the gait phase online using the \tc{robot}'s sensors. This phase can be estimated from shank \cite{Holgate2009} or thigh motion \cite{Quintero2018a}. Alternatively, Thatte \textit{et al.} recently introduced an Extended Kalman Filter (EKF) to estimate phase and its derivative (phase rate) without relying on any particular sinusoidal pattern in the sensors \cite{Thatte2019}. This EKF was also shown to yield more accurate phase estimates in non-steady locomotion when compared to conventional timing- and EMG-based controllers. Building on this result, we pursue further improvements in estimation using the Kalman Filter framework by incorporating other variations in task beyond speed.

To handle transitions between tasks, several groups have proposed solutions rooted in machine learning classifiers. Such classifiers can detect the human's intended task from patterns in the exoskeleton's sensor signals to apply the correct task-specific controller (\textit{e.g.}, stair ascent vs. level-ground walking) \cite{Tucker2015, Huang2011, Young2013, Joshi2013, Young2016, Liu2017, Hu2019, Kang2021, shepherd2022deep}. While this approach can identify discrete changes in task, it is less ideal for detecting continuous variations within a family of tasks or handling tasks outside the training data. Recently developed gait models have introduced task variables such as ground slope or stair height that continuously parameterize the instantaneous task \cite{Embry2018,Embry2020}. These variables capture more of the task's features, and can thus provide more tailored assistance to the user. Controllers based on these models have been limited to measuring the task parameter only once per stride \cite{Embry2020}---similar to the rate of phase progression before phase-based controllers were introduced. A notable exception to this paradigm is the work of Holgate \emph{et al.} \cite{Holgate2009}, which exploited the relation in the phase plane between tibia angle and angular velocity to continuously estimate gait phase and stride length. However, this relation does not hold for non-steady-state walking, nor does it extend to other joints or task variables (\textit{e.g.}, ground inclination). Recent work has also combined ambulation mode classification with continuous task variable estimates for ramp incline, step height, and walking speed \cite{Camargo2021}, but this approach uses multiple EMG electrodes, IMUs, and goniometers that may not all be available onboard an exoskeleton.

This paper introduces an EKF-based exoskeleton controller that continuously learns both the phase state (phase and phase rate) and task state (ramp and stride length) to modulate the torque profile of an assistive ankle exoskeleton in a biomimetic fashion. The EKF can indirectly estimate the gait and task parameters in real time using onboard sensors, letting the controller adapt its output quickly and in response to a continuously-varying environment. The contributions of our work include 1) introducing a new EKF phase estimator that also estimates task parameters in continuous time, 2) validating the quality of the state vector estimates using leave-one-out cross-validation based on previously-collected motion capture data of 10 able-bodied subjects walking on various inclines at various speeds, 3) validating the EKF estimates on an ankle exoskeleton \tc{used by another 10 able-bodied subjects} on an instrumented treadmill which can vary speed and inclination, and 4) validating the EKF-based controller on an ankle exoskeleton during outdoor free-walking on continuously varying surfaces (the Michigan Mars Yard and the Michigan Wavefield). These contributions to exoskeleton control enable practical, real-world usage of exoskeletons that adapt their assistance during walking at non-steady-state conditions within a continuously evolving task.

\section{Modelling and Estimating Gait}

\subsection{Gait Model}
\begin{figure}[!ht]
    \centering
    \includegraphics[width=1.0\linewidth]{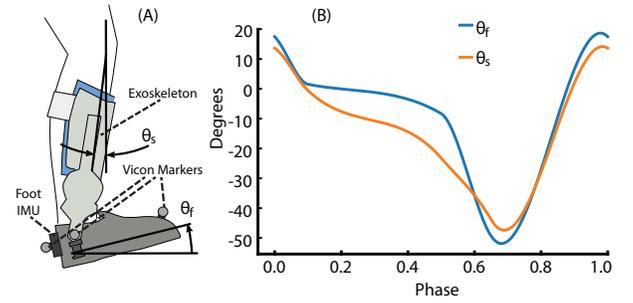}
    \caption{\label{fig:jointDefs} (A) A sagittal-plane view of the human leg. The foot angle $\theta_f$ is defined from the global horizontal, and the shank angle $\theta_s$ is defined from the global vertical. \tc{The Vicon markers used in the experiment were placed on the heel, the front of the foot (on the second metatarsal), and the ankle joint. The dedicated foot IMU was also placed on the back of the heel, while the exoskeleton system contained the IMU used for shank angle estimation.} (B) Average $\theta_f$ and $\theta_s$ profiles over the gait cycle at 1 m/s, zero incline.}
\end{figure}

Our ankle exoskeleton controller is based on a biomechanical model of gait kinematics. This model predicts global shank angle $\theta_s$, global foot angle $\theta_f$ (see Fig.~\ref{fig:jointDefs}), \tc{forward heel position $p_f$, and upward heel position $p_u$}. As input, the model takes in a gait-state vector $x$, comprising a phase (or normalized time) signal $p$, its time derivative $\dot p$, a stride length signal $l$, and a ramp angle signal $r$. The phase variable ranges from 0 to 1 and increases monotonically throughout strides, resetting at ipsilateral heel-strikes. We denote this kinematic model,
\begin{align}
    \begin{pmatrix}
    \theta_s(t) & \theta_f(t) & p_f(t) & p_u(t)
    \end{pmatrix}^T &= h_{\mathrm{gait}}(x(t)).
\end{align}

This gait model is used to infer the gait-state vector from the measurable quantities in real-time using the framework of the Extended Kalman Filter. Using the gait-state estimate, the controller then applies the corresponding bio-mechanical torque using a second model, and this allows torque profiles to vary continuously with incline angle and stride length. We use global angles and foot positions, as opposed to joint angles, because of the convenient relationship between global foot angle and ramp inclination during stance, and because they can be either measured directly or estimated with IMUs.

\subsubsection{Constrained Least-squares Regression}\label{sec:constrained_LS}
The model $h_{\mathrm{gait}}(x)$ is based on labeled training data from a 10-subject able-bodied dataset \cite{Embry2018}. \tc{This dataset contains individual stride walking data at 27 combinations of three speeds (0.8, 1, and 1.2 m/s) over nine slopes (-10 to 10 deg in increments of 2.5 deg), which were collected on a Bertec instrumented treadmill in a laboratory environment using Vicon motion capture.} Each stride features 150 samples of kinematic and kinetic data, from which we calculated phase progression over the stride. Thus the dataset provides labeled tuples of $(\theta_s(t), \theta_f(t), p_f(t), p_u(t), x(t))$ for all ($>$25,000) individual strides.

We structure $h_{\mathrm{gait}}(x)$ as 
\begin{equation}
\label{eq:gaitModelEq}
    h_{\mathrm{gait}}(x) = \phi^T R^T(x),
\end{equation}
where $\phi \in \mathbf R^{160\times4}$ is a matrix of real-valued model parameters and $R:\mathbf R^4 \mapsto \mathbf R^{1\times160}$ is a gait-state-dependent regressor row-vector.
The parameters $\phi$ are chosen to minimize the sum squared error for the equation
\begin{equation}
    \begin{pmatrix} \theta_s(t) & \theta_f(t) & p_f(t) &  p_u(t)\end{pmatrix} = R(x(t)) \phi, \label{eq:OLS}
\end{equation}
over all the sets of $(\theta_s(t), \theta_f(t), p_f(t), p_u(t), x(t))$ in the dataset. 


The regressor $R(x)$ heavily uses the Kronecker product, $\otimes$, in its construction. The Kronecker product of row-vectors $A\in\mathbf R^{1\times N}$ and $B\in\mathbf R^{1\times M}$, denoted $A\otimes B\in\mathbf R^{1\times NM}$, is the block row-vector $\large(a_1 B\ \ a_2 B\ \ \cdots\ \ a_N B\large)$. For matrices $A\in\mathbf R^{n\times N},\ B\in\mathbf R^{m\times M}$, this generalizes to \begin{equation}A\otimes B  = \begin{pmatrix}a_{11}B& a_{12}B&\cdots&a_{1N}B\\ a_{21}B & a_{22} B & \cdots& a_{2N}B\\
\vdots & \vdots & \ddots&\vdots\\a_{n1}B & a_{n2}B & \cdots& a_{nN}B\end{pmatrix}\in \mathbf R^{nm\times NM}.\end{equation}
To give a relevant example, letting $\Lambda_1(x_1) = [1,\ x_1]$ and $\Lambda_2(x_2) = [1,\ x_2]$, the combined model, $\Lambda_1(x_1)\otimes\Lambda(x_2) = [1, x_2, x_1, x_1x_2]$, represents a basis for functions which depend linearly on both $x_1$ and $x_2$.

Using a series of Kronecker products, we define the regressor
\begin{equation}
    R(x) = \Lambda_r(r)   \otimes \Lambda_l(l) \otimes \Lambda_p(p),
\end{equation}
which combines the effects of the four simpler behaviors such that the final model depends on $p$, $l$, and $r$. The components are:
\begin{itemize}
    \item The ramp angle basis $\Lambda_r:\mathbf R\mapsto \mathbf R^{1\times 2}$ is a first-order polynomial Bernstein basis \cite{bernsteinwikipedia} in ramp angle, 
        \begin{equation}
           \Lambda_r(r) = \begin{pmatrix} r & (1-r) \end{pmatrix}, \label{eq:lambda_r}
        \end{equation}
        which allows for continuous adjustment to ground slope.
    \item The stride length basis $\Lambda_l:\mathbf R\mapsto \mathbf R^{1\times 2}$ is another first-order Bernstein polynomial basis in stride length,
        \begin{equation}
           \Lambda_l(l) = \begin{pmatrix} l & (1-l) \end{pmatrix}, \label{eq:lambda_l}
        \end{equation}
        which similarly allows for kinematic changes associated with step length.
    \item Finally, the phase-polynomial basis $\Lambda_p:\mathbf R\mapsto \mathbf R^{1\times 2N}$ is a Fourier series basis of order $N$, defined as
    \begin{align}
        \Lambda_p(p) = \begin{pmatrix}
            1, \mathrm{cos}(1\cdot2\pi p), 
            \mathrm{sin}(1\cdot2\pi p), \dots\\
            \mathrm{cos}(N\cdot2\pi p),
            \mathrm{sin}(N\cdot2\pi p)
        \end{pmatrix}, \label{eq:lambda_p}
    \end{align}where $N=20$.
\end{itemize}


\begin{figure*}[!ht]
    \centering
    \includegraphics[width=1.0\linewidth]{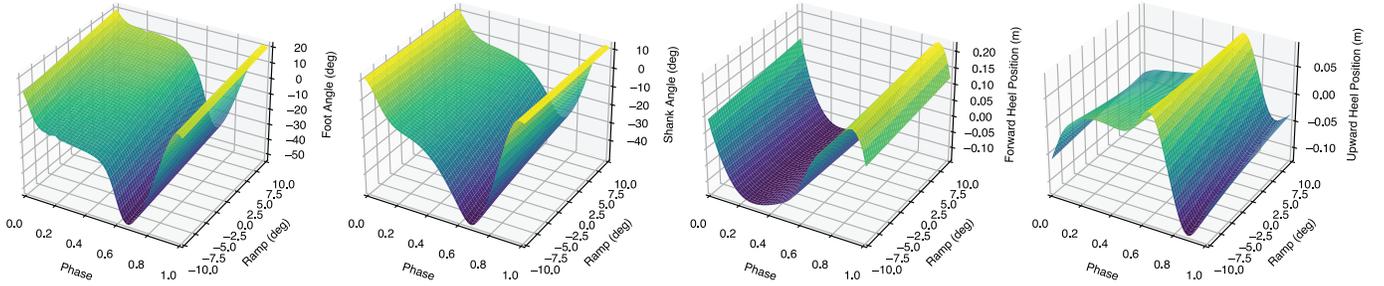}
    \caption{\label{fig:gaitModel} The regressed continuous gait models for $\theta_f$, $\theta_s$, $p_f$, and $p_u$ \tc{(foot angle, shank angle, forward heel position, and upward heel position respectively)}. As the models themselves depend on three variables ($p$, $l$, and $r$) and each produce an output, they fully reside in 4D-space and are thus difficult to express in 3D-space. In this figure, the model relation between phase and ramp is shown (with stride length constant at 1 meter). Stride length merely changes the amplitudes of the gait model.}
\end{figure*}

\subsubsection{Least-squares Constraints}
\tc{The elements of the parameter matrix $\phi$ are subject to constraints which ensure the resulting gait model $h_{\mathrm{gait}}(x)$ is well-behaved in two ways.} The constraints guarantee that the $h_{\mathrm{gait}}(x)$ function 1) predicts constant kinematics if stride length is zero (\textit{e.g.}, if the person is stationary, the measurements do not change with respect to phase), and 2) predicts the global foot angle is equal to the ramp angle when stride length is zero or when phase takes the value $0.2$.

We first define a set of two constraints to ensure constant-with-phase behavior when stride length is zero, \textit{i.e.}, when the person is standing still. The first constraint concerns the sinusoidal phase terms in the case case where stride length $l$ is zero. In this case, we expect global foot angle  $\theta_f$, global shank angle $\theta_s$, forward heel position $p_f$, and upward heel position $p_u$ to have zero coefficients for the sinusoidal terms, which we express as a matrix equality,
\begin{gather}
\overbrace{I_2 \otimes\Lambda_l(0)}^{\forall r,\ \text{if}\ l=0,} \otimes \overbrace{\begin{pmatrix}%
0 & I_{2N}
\end{pmatrix}}^{\text{sinusoid terms}}
\phi
= \overbrace{0\in \mathbf R^{4N\times4}}^{\text{are zero}}.\label{eq:zero_stride_constraint_1}
\end{gather}

To finish our pair of constraints for the case where the human is standing still ($l$=0), we consider the constant terms. In this case, the shank is vertical and the foot is aligned with the ramp, while the heel positions are assumed to be at zero position. We choose---without loss of generality---$r=0$ and $r = 10$ to express this constraint
\begin{gather}
\overbrace{\begin{pmatrix}\Lambda_r(0)\\\Lambda_r(10)\end{pmatrix}}^{\text{if}\ r=0,10} \otimes\overbrace{\Lambda_l(0)}^{\text{and}\ l=0,} \otimes \overbrace{\begin{pmatrix} 1 & 0_{1 \times 2N}\end{pmatrix}}^{\text{constant terms}}
\phi
= \overbrace{\begin{pmatrix} 0 & 0 & 0 & 0\\10&0&0&0 \end{pmatrix}}^{\text{match priors}}.
\end{gather}

Next, we constrain the model to predict that, regardless of stride length, the foot angle will be equal to $r$ at $p = 0.2$ to represent flat-foot contact. \tc{This modeling choice enforces equivalency between the foot angle and the ground incline, which allows for better estimation of incline \cite{medrano2022analysis}.}
We express this constraint on the foot using the following equality: 
\begin{gather}
\left[\overbrace{\Lambda_{\forall r}}^{\forall\ r}
\otimes \overbrace{\Lambda_{\forall l}}^{\forall\ l}
\otimes \overbrace{\Lambda_p(0.2)}^{\text{at\ }p=0.2}  \right]
\phi
\overbrace{\begin{pmatrix} 1 \\ 0 \\ 0 \\0 \end{pmatrix}}^{\text{selecting\ } \theta_f}
= \overbrace{\begin{pmatrix}0 \\ 0 \\ 10 \\ 10 \end{pmatrix} }^{\theta_f=r}, 
\end{gather}
where 
\begin{equation}
    \Lambda_{\forall r} = \begin{pmatrix} \Lambda_r(0)\\ \Lambda_r(10) \end{pmatrix},
\quad\text{and}\quad
    \Lambda_{\forall l} = \begin{pmatrix}
    \Lambda_l(0)\\ \Lambda_l(1)
    \end{pmatrix},
\end{equation}
are matrices in $\mathbf {R}^{2\times2}$ that expand the constraint to affect all values of ramp and stride length. The values $r=0$, $r=10$, $l=0$, and $l=1$ are again chosen without loss of generality to constrain the entirety of these linear functions.

\subsubsection{Complete Gait Model}
We performed the regressions for the foot and shank angle models using the constrained least-squares optimization function {\tt lsqlin} in MATLAB. The resulting models (Fig. \ref{fig:gaitModel}) not only described how the measured kinematics varied with phase, but also with ramp and stride length. \tc{Least squares regression was used to fit a Fourier series modeling gait's cyclic relationship with phase, while also allowing simpler linear fits to the other task variables that capture the salient features of gait \cite{medrano2022analysis}.}

\subsection{Biomimetic Exoskeleton Torque Profile}

\begin{figure}[!ht]
    \centering
    \includegraphics[width=0.8\linewidth]{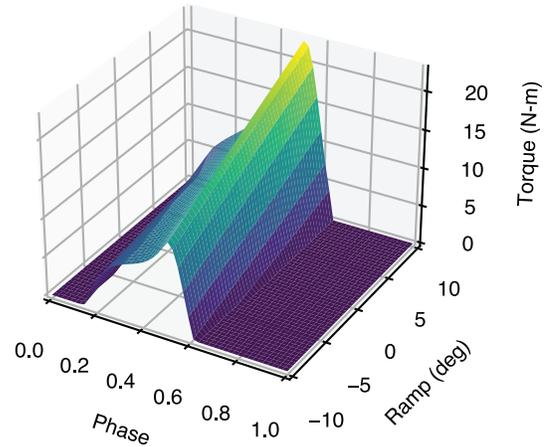}
    \caption{\label{fig:torque_profile} \tc{The regressed continuous biomimetic torque profile. Like the continuous gait model, the full torque profile depends on $p$, $l$, and $r$, and thus fully resides in 4D-space. The model relation between phase and ramp is shown (with stride length constant at 1 meter) in this figure for the purposes of visualization.}}
\end{figure}

During the live trials, the exoskeleton provided torque assistance according to a profile parameterized by phase\tc{, stride length, and incline angle}. We regressed this biomimetic torque model (Fig. \ref{fig:torque_profile}) using \tc{scaled-down biological ankle torques (by a factor of 5) from the dataset \cite{Embry2018}}. The regression was performed using a near-identical regressor structure as in (4) without the constraints on stride length and incline and with the biological ankle torque in lieu of the measured kinematics. This yielded a torque profile encoded with the same structure as the gait model in (3). \tc{As the exoskeletons used in this experiment only applied plantarflexion torque, dorsiflexion torques were floored at zero for the biomimetic torque profile.}

\subsection{Dynamic Model and State Estimator}
\subsubsection{Primary Phase EKF Model}

We use the standard EKF equations given in Appendix \ref{sec:ekf}. We define the state vector as the gait state in (2). The system process is encoded using the state transition matrix $F$:
\begin{equation}
    F = \begin{bmatrix}
   1 & \Delta t & 0 & 0 \\
   0 & 1 & 0 & 0 \\
   0 & 0 & 1 & 0 \\
   0 & 0 & 0 & 1 \\
   \end{bmatrix}\label{eq:Fdef}
\end{equation}
such that phase is updated by simple numerical integration of phase rate using the time stride $\Delta t$. 

The state covariance matrix is initialized as $P = 1\times 10^{-3} \cdot I_{4\times4}$. We define our process noise matrix $\Sigma_Q$ as a diagonal matrix $\mathrm{diag}[0, \sigma_{22}^2, \sigma_{33}^2, \sigma_{44}^2] \cdot \Delta t$, where $\sigma_{22}$, $\sigma_{33}$, and $\sigma_{44}$ are the standard deviations for $\dot p$, $l$, and $r$, respectively. Phase $p$ has no process noise since it is defined using a noiseless integration of $\dot p$. The diagonal variances act as tunable parameters that modulate EKF performance; we empirically tuned the performance and found that $\sigma_{22} = 6\times10^{-4}$, $\sigma_{33} = 9\times10^{-4}$, and $\sigma_{44} = 6\times10^{-3}$ yielded good performance with respect to phase tracking and response time. \tc{These values were held constant for all subjects.}

\subsubsection{Measurement Model}
Within the update stride of the EKF, our observation function $h(x)$ extends the directly measurable variables $\theta_f$, $\theta_s$, $p_f$ and $p_u$. To encode time-dependent measurement information, we also model the velocity of the foot $\dot\theta_f$ and shank $\dot\theta_s$. These velocities are defined using the differentiation chain rule:
\begin{equation}
    \begin{bmatrix}
        \dot\theta_f\\
        \dot\theta_s
    \end{bmatrix} = 
    \begin{bmatrix}
    \frac{\partial \theta_f}{\partial t}\\
    \frac{\partial \theta_s}{\partial t}
    \end{bmatrix} = 
    \begin{bmatrix}
    \frac{\partial \theta_f}{\partial p}\\
    \frac{\partial \theta_s}{\partial p} 
    \end{bmatrix}\dot p
\end{equation}
where $\dot p$ is the estimate of the phase rate from the prediction stride and the partial derivatives of $\theta_f$ and $\theta_s$ are available analytically. The observation function is then $h(x) = \begin{bmatrix}
\theta_f , \dot\theta_f , \theta_s , \dot\theta_s,  p_f, p_u
\end{bmatrix}^T$, where $\theta_f$, $\theta_s$, $p_f$, and $p_u$, are available from the gait model $h_{\mathrm{gait}}(x)$.

\subsubsection{Nonlinear Stride Length Transformation}

We choose to apply a nonlinear transformation to the stride length state. This transformation encodes the upper limit on a person's stride length. Furthermore, we model the smallest possible stride length as 0, which encodes backwards walking as having positive stride lengths and negative phase rates. In this transformation, the stride length is the output of an arctangent transformation \cite{VillarrealPoonawalaGregg2017TNSRE}, in which the `pseudo-stride length' $l_p$ is input. Additionally, in our gait model regression, the stride lengths were normalized by participant leg length $L$. As part of the non-linear transformation, we denormalize by participant leg length to obtain stride length $l$ in meters. The nonlinear transformation is defined as:
\begin{equation}
    l(l_p) = L \left(\frac{4}{\pi} \mathrm{atan}\left(\frac{\pi}{4} l_p\right) + 2\right).
\end{equation} 
This allows a maximum normalized stride length of 4 leg lengths and floors it at 0. $l_p$ is then the state estimated by the EKF and contained in state vector $x$. Similarly, the gait model $h_{\mathrm{gait}}(x)$ takes as input $l/L$ instead of $l$. However, for ease of communication, we refer to $x$ as containing stride length $l$ rather than its `pseudo', denormalized counterpart, and the gait model as taking the stride length input directly. To account for this change in the Jacobian $H$ in the update step of the EKF, we pre-multiply all partial derivatives with respect to $l$ by $\frac{\partial l}{\partial l_p}$.


\subsubsection{Heteroscedastic Noise Model}

EKFs generally encode measurement noise in a \emph{constant} $\Sigma_R$ matrix, which typically denotes how trustworthy the sensors used are. However, this constant model is unable to selectively change the trust in the measurements during regions of the state space where those measurements are known to be informative. For example, we expect that for phase values corresponding to flat-foot contact during locomotion, the measurements of foot angle will be highly informative for the ramp angle, given the position constraint of foot contact. To improve the performance of our phase EKF controller, we implemented a \emph{heteroscedastic} measurement noise matrix, that can continuously change the measurement noise matrix $\Sigma_R$ defined as follows: 
\begin{equation}
    \Sigma_R(p) = \Sigma_{R, \mathrm{sensor}} + \Sigma_{R, \mathrm{xsub}}(p).
\end{equation}

\begin{figure}[!ht]
    \centering
    \includegraphics[width=0.9\linewidth]{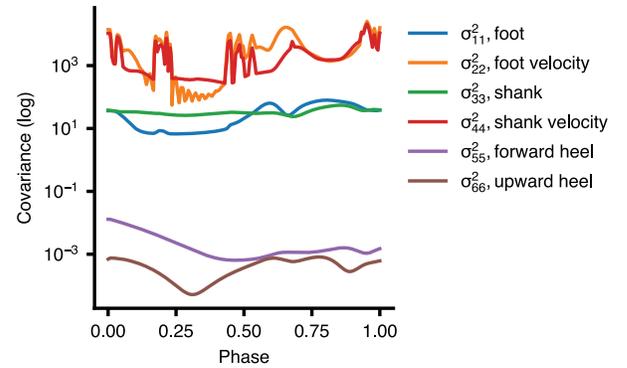}
    \caption{\label{fig:heteroscedastic} The heteroscedastic measurement noise model \tc{that modulates the measurement noise matrix} as a function of phase. Foot angle variance $\sigma_{11}$ \tc{(\textit{i.e.}, the covariance with itself)}, foot angular velocity variance $\sigma_{22}$, shank angle variance $\sigma_{33}$, shank angular velocity variance $\sigma_{44}$, forward heel position variance $\sigma_{55}$, and upward heel position variance $\sigma_{66}$, are shown. For ease of viewing, the covariances between these measurements are not shown.}
\end{figure}

In our measurement noise model, $\Sigma_{R, \mathrm{sensor}}$ is the conventional noise matrix that denotes how uncertain the sensors are, and $\Sigma_{R, \mathrm{xsub}}(p)$ represents the uncertainty present due to inter-subject gait kinematic variability (subscript $\mathrm{xsub}$ for cross-subject). $\Sigma_{R, \mathrm{xsub}}(p)$ captures the regions within the gait cycle where measurements are more informative due to lower inter-subject variability (Fig.~\ref{fig:heteroscedastic}). \tc{To determine $\Sigma_{R, \mathrm{xsub}}(p)$, we first calculated the covariance matrices of the residuals $y$ between the measurements in the prior dataset and the regressed gait model at each of the dataset's 150 phase values. These covariances incorporated the full stride data from all ten subjects in the dataset across all the dataset's conditions. In real-time, these 150 matrices were used to map the real-time phase estimate to a corresponding covariance $\Sigma_{R, \mathrm{xsub}}(p)$ via linear interpolation.}

\tc{The matrix $\Sigma_{R, \mathrm{sensor}}$ was defined as $\mathrm{diag}[\sigma_{11,r}^2, \sigma_{22,r}^2, \sigma_{33,r}^2, \sigma_{44,r}^2, \sigma_{55,r}^2, \sigma_{66,r}^2]$, with each $\sigma_{xx,r}$ representing the standard deviation for $\theta_f , \dot\theta_f, \theta_s , \dot\theta_s , p_f, p_u$, respectively. In our implementation, these values were set to 1, 10, 7, 20, 0.01, and 0.08, respectively. These different values reflected the different measurement units (\textit{e.g.}, deg vs. deg/s) and the different levels of sensor noise present in the measurements. For example, the shank angle was estimated with a secondary attitude EKF that updated at 100 Hz, whereas the foot angle was measured directly using a dedicated IMU that updated at 1 kHz, so we placed greater trust in the foot angle measurements.}

\subsection{Heelstrike-based Estimation Backup}
\tc{Depending on the process and measurement covariance matrices, it is possible that the EKF may `get lost' in its state estimation and fail to track the actual states, for example, if the filter's effective bandwidth is too slow to respond to the measurements, or if it is too fast and thus sensitive to noise.} To protect the EKF controller against getting lost in its estimation, we introduced a backup process that was permitted to reset the estimator state in the event of sub-par phase estimation. This backup system fuses a similar approach to the conventional ``Timing Based Estimation" (TBE, \cite{Malcolm2013}), which detects heel strikes and records the timestamps for each heel strike event, with the EKF framework. Our backup plan is essentially a `smaller' EKF that updates its phase rate estimates once per heel-strike, but estimates stride length and incline in real-time using the same measurements as the primary EKF.

If the backup describes a better fit to the subject's kinematics than the EKF, the backup resets the EKF state vector $x$ with its own backup state vector. The backup system computes the sum of squares of the residuals (SSR) using the residual vector $\bar y$ each stride at heel-strike. It then compares its SSR to the SSR from the phase EKF (calculated using $y$ from the EKF) over that stride. If the SSR from the backup is sufficiently smaller than the SSR from the EKF, \textit{i.e.}, the EKF is performing poorly, then the backup overrides the EKF and places it back on course. We tuned the backup EKF to only activate once every 100 steps when the SSRs are of comparable magnitude.

\section{Study Methods}

\tc{We perform data-driven simulations and hardware experiments to} address the following hypotheses:
H1) Our EKF-based controller has a significantly lower phase RMS error compared to phase estimates from a HS-to-HS timing based controller (state-of-the-art) in a leave-one-out cross validation;  
H2) The inclusion of \tc{task variables} significantly improves the phase RMS error in a leave-one-out cross validation; and 
H3) Our real-time EKF estimate of phase has an RMS error less than that of state-of-the-art timing based estimators in the presence of actuator torques. We also present the result of the EKF working in a practical outdoors setting with exoskeleton actuation. All human participants gave written, informed consent with approval from the University of Michigan Institutional Review Board (Study ID: HUM00158854).

\subsection{EKF Simulation}

We cross-validated our EKF \emph{in silico} using the same walking dataset \tc{mentioned in Sec. \ref{sec:constrained_LS} (N=10 subjects \cite{Embry2018}), which includes trials at three speeds (0.8, 1, and 1.2 m/s) over nine slopes between $-10$ and $10$ deg. }The dataset contains walking data indexed by walking stride, which was used for training the gait model $h_{\mathrm{gait}}(x)$. This stride data was concatenated to form a continuous walking sequence, which was then input into our EKF to analyze performance with realistic locomotion. The complete source code for our simulation is available in a ready-to-run computation capsule format through CodeOcean \cite{MedranoThomasGreggRouse2022CodeOcean}.

\subsection{Cross-Validation of Simulated Phase Estimate Quality (H1)}\label{sec:h1}

To identify the ability of our EKF to adapt to new subjects while accurately estimating phase, we performed leave-one-out cross-validation on our EKF-based controller using the dataset, where ground truth phase was calculated using the normalized time between heel-strikes. For each subject, this cross-validation trained a new gait model using the individual stride data from the remaining nine subjects. \tc{Similarly, we also trained a new heteroscedastic model for each subject with the residual covariance data from the remaining nine subjects.} We then used the concatenated stride walking data from the subject not included in the training set as the input to our EKF simulator, and computed the phase root-mean-squared-error (RMSE) of our EKF relative to the ground truth phase measurements for each stride in the dataset for that subject. To determine the improvement produced by our EKF controller, we compared the stride-wise phase errors from the EKF against the phase errors from a simulated TBE, which predicts the current phase in real-time using the normalized timings of previous heel strikes. This common method is causal whereas ground truth phase is non-causal. The TBE estimator was assumed to perfectly detect all heel-strikes from the ground truth dataset (an assumption that may not hold during practical implementations). Using a \tc{two-tailed}, paired t-test, we compared differences in the stride-wise phase RMSE between the EKF and TBE approaches. We also computed the stride-wise stride length and incline RMSE relative to the ground truth values from the dataset. \tc{To assess how the resulting control torques were affected by the gait state estimates from the EKF, we computed the stride-wise RMSE between the torque profile resulting from the EKF state estimates and the torque profile resulting from the ground truth states.}

\subsection{\tc{Cross-Validation without Task Variable Estimation (H2)}}\label{sec:h2}

\tc{In a separate leave-one-out cross-validation simulation, we assess the improvement to phase estimation quality that comes from the inclusion of the stride length and incline task variables in the state vector. This allows for a more direct comparison to the EKF featured by Thatte \textit{et al.} \cite{Thatte2019}, which only used phase variables in their filter. To effectively eliminate the stride length and ramp state variables, elements of the process noise matrix $\Sigma_Q$ and covariance matrix $P$ corresponding to these variables were initialized to extremely low values on the order of $1e{-12}$, which prevented the estimates from changing in real time. These values were not simply set to zero in order to avoid numerical issues in the real-time computation of the state estimate. We then computed the phase RMSEs for each stride using the phase estimates of this limited EKF and determined the significance of including the task variables by comparing these errors to those from the proposed EKF using a two-tailed, paired t-test.}

\subsection{Hardware Setup}
We implemented our phase EKF on an ankle exoskeleton system (ExoBoot, Dephy, Inc., Maynard, MA) for our human participant experiments. We used separate \tc{sensors} to measure each of the global angles due to the compliance of the exoskeleton's structure. We estimated the global shank angle using a custom attitude EKF \cite{julier2004unscented} that ran concurrently with our phase EKF. The attitude EKF used readings from the onboard IMU (TDK Invensense MPU-9250, San Jose) mounted above the ankle joint of the exoskeleton. To measure global foot angle, we used an external IMU (3DM-GX5-25, LORD Microstrain, Williston) placed on the boot of the exoskeleton. This IMU estimated global orientation from which we obtained the global foot angle (\textit{i.e.}, pitch angle). \tc{To account for inter-subject differences in exoskeleton position and resting shank angle, we calibrated both of these sensors prior to each trial by having the subjects stand still and recording the average angular measurements from these sensors, which were then subtracted from the live measurements.} For the angular velocity measurements, we measured the shank angular velocity directly using the exoskeleton's gyroscope readings, and measured foot angular velocity using the foot IMU's gyroscope readings. \tc{To obtain the heel position signals, we applied a second-order linear filter to the foot IMU's accelerometer signals, which combined a double-integrator with a high-pass filter to convert forward and vertical acceleration signals into forward and vertical heel positions while also high-pass filtering away the effects of integration drift and gravity}.

For the indoor hardware experiment, \tc{ten} able-bodied, human subjects walked on an instrumented Bertec treadmill (Bertec, Columbus OH) with variable belt speed and inclination. The outdoor experiments were conducted with a single able-bodied human subject on both the University of Michigan Mars Yard (GPS: 42.29464, -83.70898), which demonstrates an unstructured outdoor environment, and the University of Michigan Wavefield (GPS: 42.29323, -83.71168), which provides a practical stress test on extremely unsteady terrain.
\subsection{Treadmill Experiment (H3)}\label{sec:h3}

We characterized our controller's performance during a controlled treadmill experiment. \tc{In this experiment, ten able-bodied participants (5 male, 5 female, age: $24.2\pm2.15$ years, height: $1.75 \pm0.08$ m, weight: $74.4 \pm 14.1$ kg) walked with the EKF-controlled exoskeleton for three segments.} In the first segment, participants continuously walked on level ground for twenty seconds at 1.2 m/s, followed by twenty seconds at 0.8 m/s, followed by a return to 1.2 m/s for twenty seconds, then finally for 0.8 m/s for a final twenty seconds. In the second segment, the participants walked on the treadmill for 10 seconds at 1 m/s, then the treadmill inclined over a period of one minute and ten seconds to a maximum of 10 deg while the participant continued walking. The third segment was identical to the second, except that instead of the treadmill inclining to 10 deg, it declined to -10 deg. \tc{The exoskeleton applied the adaptive biomimetic torque throughout the trials via PID current control to apply the desired torque at the ankle.} 

Participants were instrumented with Vicon markers to capture their kinematics, which were then used to establish ground truth estimates for phase, phase rate, stride length, and inclination, as well as for heel-strike events. \tc{Ground truth phase rate for each stride was estimated by taking the inverse of that stride's duration; ground truth phase was then calculated by integrating this rate over the duration of the stride beginning at heel-strike. Ground truth stride length was then calculated by dividing the known speed by the phase rate, while ground truth inclination was measured as the inclination of the treadmill.} The Vicon experimental system featured 27 cameras and collected kinematic data at 100 Hz. We computed the errors and RMSEs for both EKF phase, phase rate, stride length and incline. We then compared the stride-wise phase RMSEs to those from a state-of-the-art TBE approach which thresholded the signals from the onboard device sensors to detect heel-strikes (similar to the approach from \cite{Mooney2014}); the thresholds were tuned to correctly detect heel-strikes across the walking tasks in this treadmill trial. We compared the two controllers using the same t-test as with the previous hypotheses.

\subsection{Real-world Experiments}
Finally, we characterized the ability of the EKF-based controller to adapt its gait state estimates and torque assistance in a highly irregular outdoor environment. One participant from H3 walked with the EKF-controlled exoskeleton on both the Michigan Mars Yard, which features a steep hill (approximately $20$ deg) covered in small rocks, and the Michigan Wavefield, which features a sinusoidal hill pattern of inclines and declines (from approximately $30$ to $-30$ deg) and has been previously used to test the stability of bipedal robots \cite{da2017supervised}. These locations were chosen to `stress-test' the EKF with non-steady-state conditions outside its training dataset. During these tests, the process noise matrix $\Sigma_Q$ was tuned to have $\sigma_{22} = 1e{-3}$, $\sigma_{33} = 2e{-3}$, and $\sigma_{44} = 5e{-2}$ for a faster filter response. At the Mars Yard, the participant walked in an alternating `slow-fast-slow' speed pattern on level ground both prior to and after ascending/descending the hill. At the Wavefield, the participant performed the same alternating speed pattern before and after walking on the rapidly changing, sinusoidal hills. The separation of the speed and slope changes during these outdoor trials was done to isolate the EKF's ability to track each task variable independently. We recorded HD videos of these trial and compared the recorded events to the EKF's estimates of the gait state to assess the adaptations to the terrain features of the Mars Yard and Wavefield. To obtain ground truth phase values, we analyzed the video to determine the timings of heel-strike events, and interpolated phase values between them (assuming a constant phase rate). For this experiment, we compared our EKF against a TBE identical to the one in \tc{Sec.~\ref{sec:h1}} (recall TBE is causal whereas ground truth is not). We performed pairwise t-tests comparing the phase RMSEs calculated over each stride between the EKF and this TBE.

\begin{figure*}[!t]
\centering
{\footnotesize
\includegraphics[width=2.0\columnwidth]{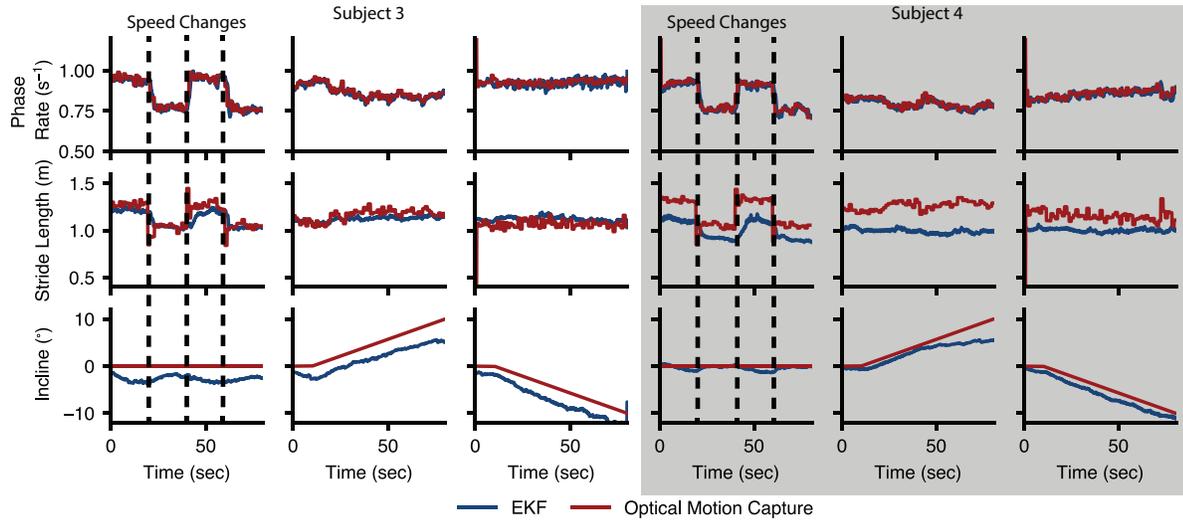}}
\vspace{-1mm}
\caption{\label{fig:treadmillTrials} State estimates from the treadmill trials from two representative participants. Phase rate is shown in the first row, stride length in the second, and incline in the third. For clarity, the periods where the speed changed in the first trial are delineated by the black dashed lines. Overall, the EKF had excellent tracking of phase rate (and thus, phase), as evidenced in these representative trials. The EKF was responsive enough to correctly estimate Subject AB03's stride length but tended to underestimate this state for some participants (such as Subject AB04), in part due to individual differences from the trained gait model. Finally, the EKF adequately tracked ground incline, although it occasionally exhibited a negative bias due to changes in gait caused by the exoskeleton assistance (more obvious in Subject AB03).
}
\end{figure*}
\section{Results}

\begin{table*}[]
\centering
\caption{Gait State RMSEs for All Subjects in Treadmill Experiment}
\label{tab:treadmill_data}
\begin{tabular}{ccccccc}
\hline
Participant                        & \begin{tabular}[c]{@{}c@{}}EKF Phase \\ RMSE (\%)\end{tabular} & \begin{tabular}[c]{@{}c@{}}EKF Phase Rate \\ RMSE (\%)\end{tabular} & \begin{tabular}[c]{@{}c@{}}Stride Length \\ RMSE (m)\end{tabular} & \begin{tabular}[c]{@{}c@{}}Incline \\ RMSE ($^{\circ}$)\end{tabular} & \begin{tabular}[c]{@{}c@{}}Desired Torque \\ RMSE (N-m)\end{tabular} & \begin{tabular}[c]{@{}c@{}}TBE Phase \\ RMSE (\%)\end{tabular} \\ \hline
AB01                               & 4.9                                                            & 1.5                                                                 & 0.22                                                              & 1.0                                                                  & 2.9                                                                  & 5.7                                                            \\
AB02                               & 4.1                                                            & 1.8                                                                 & 0.15                                                              & 2.7                                                                  & 3.3                                                                  & 5.1                                                            \\
AB03                               & 4.7                                                            & 2.4                                                                 & 0.07                                                              & 2.8                                                                  & 2.6                                                                  & 5.5                                                            \\
AB04                               & 3.3                                                            & 2.7                                                                 & 0.21                                                              & 1.2                                                                  & 2.4                                                                  & 4.9                                                            \\
AB05                               & 5.6                                                            & 1.9                                                                 & 0.25                                                              & 1.8                                                                  & 3.5                                                                  & 5.6                                                            \\
AB06                               & 2.7                                                            & 2.0                                                                 & 0.28                                                              & 3.1                                                                  & 2.5                                                                  & 3.4                                                            \\
AB07                               & 11.0                                                           & 3.0                                                                 & 0.08                                                              & 2.4                                                                  & 4.9                                                                  & 12.5                                                           \\
AB08                               & 5.4                                                            & 1.5                                                                 & 0.10                                                              & 1.7                                                                  & 2.9                                                                  & 5.2                                                            \\
AB09                               & 2.8                                                            & 1.9                                                                 & 0.12                                                              & 2.0                                                                  & 2.2                                                                  & 3.5                                                            \\
AB10                               & 3.2                                                            & 1.8                                                                 & 0.06                                                              & 5.6                                                                  & 2.8                                                                  & 3.8                                                            \\ \hline
\textbf{Mean (Standard Deviation)} & 4.8 (2.4)                                                      & 2.1 (0.5)                                                           & 0.15 (0.08)                                                       & 2.4 (1.3)                                                            & 3.0 (0.76)                                                           & 5.5 (2.6)                                                      \\ \hline
\end{tabular}
\end{table*}

\subsection{{Cross-Validations of EKF Performance (H1 and H2)}}

\begin{figure*}[!t]
\centering
{\footnotesize
\includegraphics[width=2.0\columnwidth]{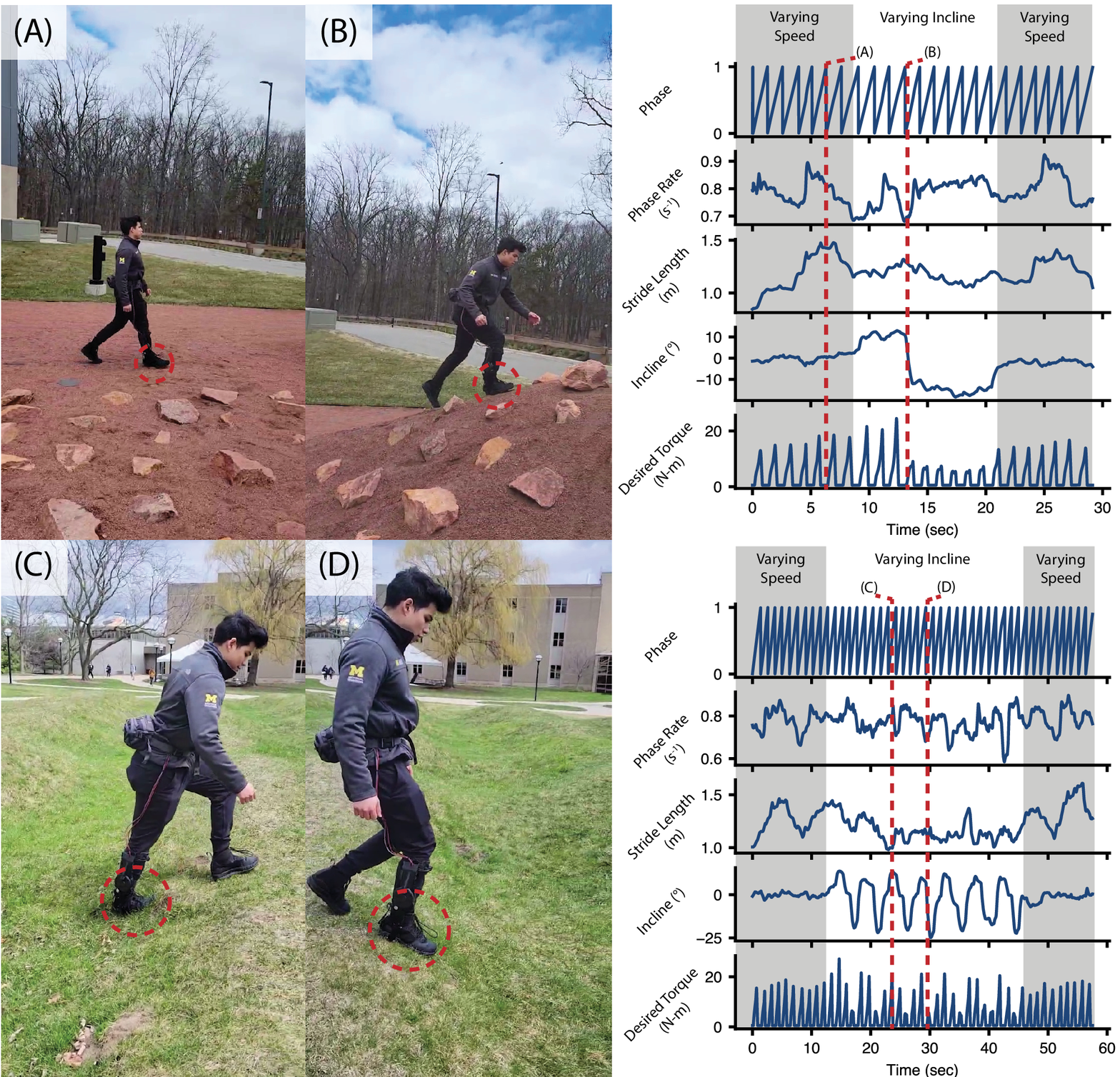}}
\caption{\label{fig:ekf_stress_test} Stills from the analyzed videos taken during the Mars Yard and Wavefield trials, with a focus on the EKF's incline estimation performance. The incline estimation is primarily driven by the participant's foot angle on the outdoor slopes (red dashed circles). (A) The subject walking on level ground on the Mars Yard during a period of varying speeds, with the EKF correctly updating its estimate. (B) The subject walking up the Mars Yard, in which the EKF correctly increases its incline estimate. (C) The subject walking uphill on the Wavefield (D) The subject going down a decline on the Wavefield.}
\end{figure*}

Our results support Hypothesis 1---that the EKF significantly outperforms TBE during an \textit{in silico} leave-one-out cross validation. \tc{In terms of \% of stride cycle progression}, the inter-subject phase RMSE for the EKF ($1.6 \pm 0.31\%$) was significantly lower ($P<0.001$) than the inter-subject RMSE for the TBE estimator ($2.1 \pm 0.1\%$). \tc{Additionally, our EKF predicted stride length with an inter-subject RMSE of $0.10 \pm 0.02$ m, ramp inclination with an inter-subject RMSE of $1.5\pm 0.23$ deg, and desired torque from our biomimetic torque profile with an inter-subject RMSE of $1.2 \pm 0.18$ Nm. These RMSEs are likely due to inter-person variability in locomotion, which have been established to cause errors in kinematic predictions \cite{Embry2020}}. 

\tc{The second cross-validation supported Hypothesis 2---that including task variable estimation in the EKF significantly improves phase estimation when compared to an EKF that does not feature task variable estimation (\textit{e.g.}, \cite{Thatte2019}). The phase RMSE for the No-Task condition ($2.0 \pm 0.36\%$) was significantly higher than the phase RMSE for the proposed EKF condition ($1.6 \pm 0.31\%$, $P<0.001$).}

\subsection{Treadmill Experiment (H3)}

The treadmill test supported Hypothesis 3---that the EKF significantly outperforms TBE while applying torques. \tc{The EKF (phase RMSE: $4.8 \pm 2.4\%$, Table ~\ref{tab:treadmill_data}) resulted in significantly lower ($P<0.002$) phase errors than the conventional TBE approach (phase RMSE: $5.5 \pm 2.6\%$). The EKF captured the pulse-like change in phase rate during the fast 1.2 m/s and slow 0.8 m/s sections of the trials, as shown in representative trials (Fig.~\ref{fig:treadmillTrials}). The EKF simultaneously provided live estimates of both stride length and inclination (cross-subject average stride length RMSE $0.15 \pm 0.08$ m, incline RMSE $2.4 \pm 1.3$ deg).} The EKF clearly responded to the changes in ground inclination, although it often exhibited negative biases, as exemplified by the results from Subject AB04 in Fig.~\ref{fig:treadmillTrials}. The EKF underestimated changes in stride length, although with notable differences in tracking quality across subjects (Fig.~\ref{fig:treadmillTrials}). \tc{Additionally, the desired torque profile RMSE, which encodes the gait-state RMSEs, was only $3.0 \pm 0.76$ Nm compared to the $\sim$20~Nm peak magnitude of the torque profiles. From our torque profile, the average (across phase) change in torque from a stride length change of $0.15$ m (the average stride length RMSE across subjects) was only $0.12$ Nm.}


\subsection{Real-world Experiments}

In the Mars Yard trial, where the subject alternated speed and incline with exoskeleton assistance, the EKF was comparable to the conventional TBE ($3.8 \pm 2.0\%$ for the former, $4.4 \pm 3.1\%$ for the latter, $P = 0.2$). However, unlike the TBE, the EKF state estimates tracked both the changing speeds prior to ascending the Hill and the changing ground slope from the Hill itself. Two stills that correspond to notable events from the trial are shown in Fig. \ref{fig:ekf_stress_test}A and B; in particular, the position of the subject's right foot is useful to demonstrate that the EKF is updating its incline estimate correctly. On the Wavefield, the EKF significantly outperformed the conventional TBE ($4.8 \pm 2.7\%$ for the former, $7.3 \pm 8.3\%$ for the latter, $P = 0.02$). The extreme conditions on the Wavefield (Fig. \ref{fig:ekf_stress_test}C and D) rendered the steady-state assumptions of the TBE a poor fit for the gait estimation task; in contrast, the EKF was able to track both phase and the rapidly changing ground incline of the Wavefield. The EKF's adaptation of its state estimates is also reflected in the assistance from the exoskeleton. The biomimetic torque profile broadly increased the magnitude of its torque assistance when ground inclination was positive, and decreased when it was negative. This trend is reflected in the torques shown in Fig. \ref{fig:ekf_stress_test}. A supplemental video of the real-world experiment is available for download.



\section{Discussion}

As expected, our EKF phase estimator outperformed the TBE estimator, with significantly lower phase RMSEs in the \emph{in silico} steady-state treadmill trials. Fundamentally, the EKF observes the behavior between heel-strikes, allowing it to better predict the phase variability that accompanies natural human walking, even in this steady-state test with a constant belt speed\tc{, within each tested condition}.

\tc{The presence of real-time task variable estimation within the EKF also significantly improved phase estimation. This is likely due to these task variables accounting for part of the variation from the kinematic sensor measurements, and therefore reducing the prediction errors. Without estimation of stride length or inclination, modeling error increased, resulting in greater variance in the phase estimation signal. Hence, the task variable states could improve phase estimation during variable-speed and variable-incline walking compared to the phase EKF presented in \cite{Thatte2019}.}

In the treadmill experiment, the EKF estimator outperformed the conventional TBE in addition to providing estimates of task variables in real-time. The EKF consistently detected heel-strikes with more accuracy than the conventional approach. \tc{Our EKF's phase RMSE (cross-subject average $4.8 \pm 2.4\%$)} is comparable to recent online phase estimation performance achieved in hip exoskeletons \cite{Kang2021}, which used machine learning to estimate phase for subjects walking through circuits that featured different ramps (phase RMSE $5.04\%$\footnote{Note that comparing RMSE across distinct experiments is not a direct comparison of methods. We also feature a similar RMSE to that of another machine learning-based estimator that was implemented on the same ankle exoskeleton hardware \cite{shepherd2022deep} ($3.9$\% stance phase RMSE).}). Additionally, our phase RMSE is near the perceptual threshold in timing error, suggesting that subjects may be unlikely to detect any errors in torque profile assistance due to the EKF's phase error \cite{peng2022actuation}. \tc{The EKF's mean incline RMSE (cross-subject average $2.4 \pm 1.3$ deg)} was also comparable to recent results from an offline ML-based sensor fusion approach \cite{Camargo2021} (absolute incline error $2.3 \pm 0.86$ deg).

In the real-world tests, and in particular during the Wavefield stress test, the EKF estimator had a significant advantage over the conventional TBE as in the treadmill trial. The conventional TBE featured its poorest performance during the Wavefield trial (up to 7.3\% from 2.1\% RMSE in the steady-state cross-validation trials), due to the highly variable nature of the terrain and the non-steady-state steps taken by the participant. The EKF, in contrast, only increased to 4.8\% phase RMSE \tc{from the steady-state cross validation phase RMSE of 1.6\%}. \tc{These results are comparable to a recent study \cite{qian2022predictive} (phase RMSE: 4.12\%) which estimated phase using a combination of IMUs and cameras to control a hip exoskeleton during stair and ramp ascent and descent.}


Using an EKF confers several advantages for our adaptive torque controller. Because our simple gait model predicts angles and velocities in a fundamentally structured way, we are able to encode a large, continuous class of locomotion tasks. The EKF equations are intuitive to understand, simple to implement on hardware, and computationally lightweight, while still yielding comparable phase estimation performance to more complex machine learning approaches. Furthermore, the data-driven model component at the core of our controller may be easier to debug than so-called `black-box' models that directly estimate gait parameters or control commands from sensor data. \tc{Similarly, it would be straightforward to substitute another gait model (for example, a Gaussian Process as in \cite{Thatte2019} or a neural network as in \cite{Camargo2021}) within our EKF depending on the preference of the experimenter, as the only requirement is that it take in the state vector as inputs and produce the kinematics as outputs.}

\forceblack{Our EKF-based controller provides a solution to the challenge of estimating the state of an exoskeleton during practical trials. This challenge has been avoided in prior exoskeleton control work \cite{Mooney2014b,Zhang2017,Ding2018,Panizzolo2019} by testing in controlled steady-state conditions that allow TBE approaches to suffice. However, these conditions are unrepresentative of the real world. The performance of our controller throughout the extreme conditions of the Wavefield test demonstrates that it can provide phase estimates throughout non-steady-state conditions, which could potentially allow exoskeleton research to be translated out of the laboratory. Additionally, our explicit gait model is agnostic to the choice of exoskeleton hardware, so our estimator could serve as a `plug-and-play' solution for researchers seeking to take their devices into the real world.}


\forceblack{Our EKF can also scale well to additional sensors. In our current implementation, we use four sensors (foot and shank angles, along with their derivatives, and forward and upward heel position), but it would be straight-forward to implement other sensors, such as instrumented insoles, to further improve phase estimation. We can use the current dataset to regress relationships between these sensors' measurements and our gait-state and simply extend the measurement vector in our EKF. }\tc{We expect that this will further improve phase estimation due to the new information available from these sensors, motivating further study into the impact of different sensors (\textit{e.g.}, \cite{medrano2022analysis}) as well as different model-fitting techniques to prevent overfitting (which we avoided by implementing a relatively simple linear model structure relative to the complexity of the dataset).}


\begin{figure}[!t]
\centering
{\footnotesize
\includegraphics[width=1.0\columnwidth]{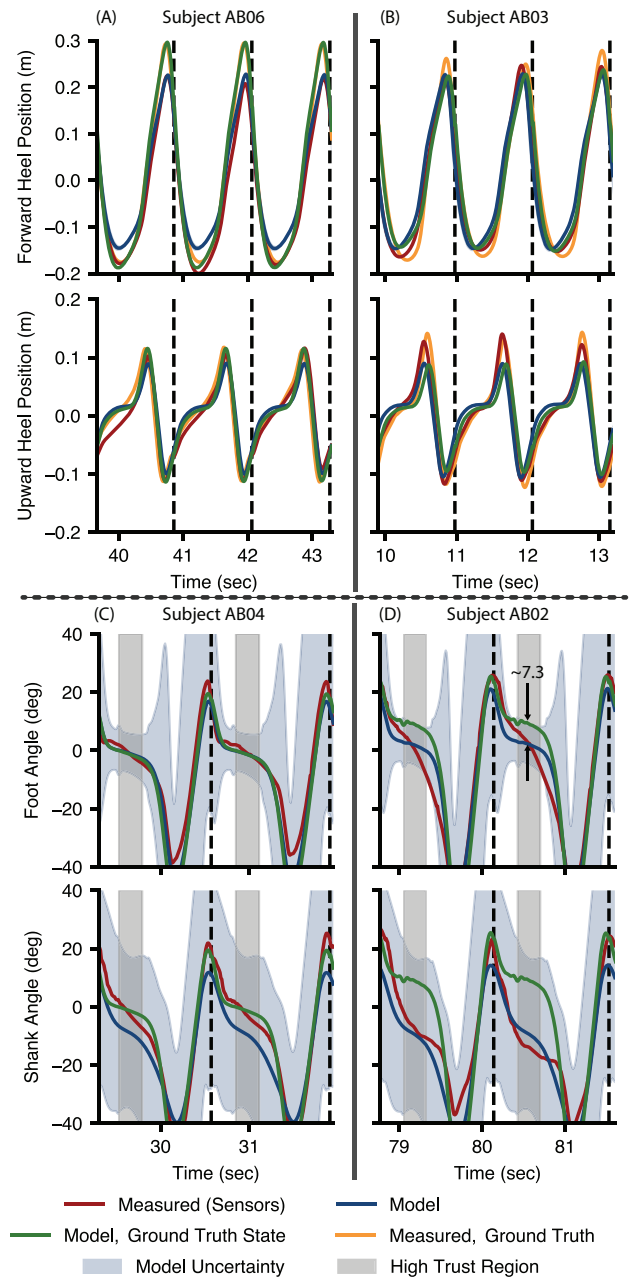}}
\caption{\label{fig:exo_discrepancies} (A) \tc{Subject AB06's forward and upward heel position kinematics during the treadmill trials, as an example of how inter-person gait variability can lead to errors in state estimation. Based on the subject's kinematics as measured by the sensors (red), the EKF gait model predicts heel positions (solid blue) that imply a stride length state estimate below the ground truth measured by Vicon (orange). The gait model predictions using the ground truth state from Vicon (green) would more closely track the actual measurements from the sensors for Subject AB06. (B) Subject AB03's forward and upward heel kinematics during the treadmill trials better match the general gait model's predictions of heel position, leading to more accurate stride length estimates. (C) Subject AB04's foot and shank angle measurements exhibit a clear flat-foot contact period in early stance during regions of high model trust (gray shade) and thus result in accurate incline estimates; trust is indicated by the width of the confidence region (blue shade) about the model estimates. (D) Subject AB02's foot and shank angle measurements exhibit a less-defined flat-foot contact period where the noise model expects high confidence in the measurement, resulting in persistent negative biases in the incline estimate.}}
\end{figure}

\forceblack{To further improve our estimates in future work, we would first target the inter-person kinematic variability that increases the variance of not only the estimated phase, but also the remaining state vector elements. }\tc{Prior work has demonstrated that inter-person variability can cause large prediction errors in subject-independent kinematic models \cite{Embry2020}. We suspect this is responsible for the large stride length RMSEs experienced by some subjects, such as Subject AB06 (0.26 m RMSE, Tab.~\ref{tab:treadmill_data}), and for the smaller RMSE in stride length ($0.10 \pm 0.02$ m) in the simulated walking cross-validation.} \tc{The EKF's estimates of stride length are primarily informed by the amplitude of the forward and upward heel position measurements. When analyzing Subject AB06's forward and upward heel positions (Fig.~\ref{fig:exo_discrepancies}A), there is a large discrepancy between the predicted position from the EKF's live gait model (blue), and the predicted position from the gait model when using the ground truth states from Vicon as inputs (green). For Subject AB06, the stride length that produced the best-matching predictions was smaller than the ground truth stride length. This is likely because the actual stride length is known with more certainty later in swing as the stride nears completion; during this region of phase in swing, the position measurements obtained by integrating the accelerometer readings from our IMU sensor (red) are smaller in amplitude than the actual positions measured by the Vicon signals (orange). The EKF sees these smaller amplitudes and estimates a smaller stride length, as it fits the measurements better. The discrepancy between the sensor measurements and ground truth measurements is far less pronounced earlier in stride, indicating the the integration was more accurate during this region. Our filter was designed to produce accurate integration of the forward and upward heel positions based on the generalized gait model. Subject AB06 simply walks differently later in the gait cycle than what our filter was designed to estimate and our model was designed to predict, in a way that the general gait model cannot capture with accurate state estimates. Notably, this only resulted in a desired torque RMSE of 2.6 Nm for Subject AB06, as the other gait state variables, in particular phase, were still estimated well. In contrast, Subject AB03 (Fig.~\ref{fig:exo_discrepancies}B) walks nearly identically to the general gait model prediction, and thus their state} \tc{estimates (in particular stride length) are more accurate. Simple gait personalization techniques have been shown to significantly reduce the error in gait models over a continuous range of tasks \cite{Reznick2020}, so we} \tc{suspect that the variance of estimated phase could be reduced with such a personalized gait model. }




\forceblack{Unmodeled dynamics can also negatively impact our EKF performance. In particular, the exoskeleton's actuation imposes a disturbance on the system through the physical deflection of the device's IMU, which can lead to incorrect measurements of the shank angle. Our gait and heteroscedastic noise models assume that the measured shank angles will take on a specific, average profile, but actuation causes the measured shank profiles to be fundamentally different than what our models expect. In our implementation, we accounted for this by increasing the shank measurement covariance so that the EKF placed less trust in this signal. While this made the controller more robust to the disturbance caused by the exoskeleton torque, it also reduced the bandwidth of the controller, which otherwise could have been tuned to more quickly detect changes in gait.}

\forceblack{Furthermore, the presence of exoskeleton assistance may have caused some participants to walk differently than how our gait model predicted, which would also lead to incorrect EKF behavior. For example, our gait model predicts that foot angle during early stance is relatively stationary and equal to the ground incline angle. Consequently, our noise model trusts the foot measurements during these regions of stance (Fig.~\ref{fig:exo_discrepancies}, gray shade) as highly informative for estimating incline (Fig.~\ref{fig:exo_discrepancies}, narrower blue bands around the predicted foot measurement during early stance). Some participants (Fig.~\ref{fig:exo_discrepancies}C), walked similarly to this prediction while experiencing powered assistance, which led to more accurate incline estimates.} \tc{However, other participants (Fig.~\ref{fig:exo_discrepancies}D) notably had foot angle measurements that sloped downward more steeply during early stance and had a less-defined foot-contact region of stance, which was also reflected in a steeper shank angle. We suspect this is due to these participants allowing the exoskeleton torque to begin lifting their heel off earlier in stance. During the region of stance where the EKF expected 1) the foot measurement to be most informative about incline, and 2) the foot angle to be equal to the ground incline, the presence of exoskeleton assistance caused the foot measurement to have a negative value relative to the ground inclination. For Subject AB02 in Fig.~\ref{fig:exo_discrepancies}D, this negative bias was as high as 7.3 deg. This was reflected in the ramp estimates of the EKF, which undershot the true ground incline by roughly this negative bias.} \forceblack{Here again, individualized gait models may be a potential solution, as they can capture not only how every person walks differently, but also how each person uniquely adapts their gait to exoskeleton assistance.}


\forceblack{Because the dataset used to regress our gait model only contained steady walking data, other tasks such as running or start/stopping are not explicitly modeled. While in theory the presented EKF can account for sudden stops by estimating $\dot p$ as zero, we believe the estimator will benefit from training with data including such gait transitions (\textit{e.g.}, \cite{Reznick2020a}). As datasets grow to include more behaviors, our intention is to extend our continuous gait model with new task variables representing these other dimensions of human locomotion.}


\section{Conclusion}

\forceblack{We developed an exoskeleton torque controller based on an EKF which estimates phase, phase rate, stride length, and ramp in real time. This controller yields significantly reduced phase estimation errors compared to the state of the art. Furthermore, this controller improves upon the state of the art by allowing the assistive torque profile to adapt in real time in response to the state estimates. To the authors' knowledge, we are the first to estimate the gait phase variable along with stride length and ground inclination in real time and throughout an outdoor non-steady-state locomotion task \tc{on difficult terrains}. This result represents a meaningful milestone for practical exoskeleton control and usage outside the laboratory.}

\section*{Acknowledgements}
\forceblack{The authors would like to thank Emma Reznick for providing invaluable training in using the Vicon motion capture system.}

\appendices
\section{Extended Kalman Filter Implementation}
\label{sec:ekf}
\forceblack{Starting from the state estimate at the previous time $\hat x_{k-1|k-1}$ and the previous state covariance estimate $P_{k-1|k-1}$, our EKF implementation computes the current state estimate update using measurement $z_k$, dynamic model $f(\cdot)$, and measurement model $h^\prime(\cdot)$. The process involves two steps. First, we propagate the (conveniently linear) dynamics from \eqref{eq:Fdef} across the time step,
\begin{align}
    \hat x_{k|k-1} &= F\hat x_{k-1|k-1},\nonumber\\
    P_{k|k-1} &= F P_{k-1|k-1} F^T + \Sigma_Q.\nonumber
\end{align}
We then correct the estimate based on the measurement $z_k$ as
\begin{align}
    \hat x_{k|k} &= \hat x_{k|k-1} + K_k(z_k - h(\hat x_{k|k-1})),\nonumber\\
    P_{k|k} &= P_{k|k-1}-K_k H_k P_{k|k-1},\nonumber\\
    \text{where\ \ }K_k &= P_{k|k-1}H_k^T\left[H_k P_{k|k-1} H_k^T + \Sigma_{R_k}(\hat{p}_{k|k-1})\right]^{-1},\nonumber\\
    \text{and\ \ }H_k &= \frac{\partial h^\prime}{\partial x}|_{\hat x_{k|k-1}} \begin{pmatrix}1&0&0&0\\0&1&0&0\\0&0&\frac{dl}{dl_p}&0\\0&0&0&1\end{pmatrix}.\nonumber
\end{align}
}

\bibliographystyle{IEEEtran}
\forceblack{
\bibliography{IEEEtran}

\begin{thebibliography}{10}
\providecommand{\url}[1]{#1}
\csname url@samestyle\endcsname
\providecommand{\newblock}{\relax}
\providecommand{\bibinfo}[2]{#2}
\providecommand{\BIBentrySTDinterwordspacing}{\spaceskip=0pt\relax}
\providecommand{\BIBentryALTinterwordstretchfactor}{4}
\providecommand{\BIBentryALTinterwordspacing}{\spaceskip=\fontdimen2\font plus
\BIBentryALTinterwordstretchfactor\fontdimen3\font minus
  \fontdimen4\font\relax}
\providecommand{\BIBforeignlanguage}[2]{{%
\expandafter\ifx\csname l@#1\endcsname\relax
\typeout{** WARNING: IEEEtran.bst: No hyphenation pattern has been}%
\typeout{** loaded for the language `#1'. Using the pattern for}%
\typeout{** the default language instead.}%
\else
\language=\csname l@#1\endcsname
\fi
#2}}
\providecommand{\BIBdecl}{\relax}
\BIBdecl

\bibitem{Mooney2014b}
L.~M. Mooney, E.~J. Rouse, and H.~M. Herr, ``{Autonomous exoskeleton reduces
  metabolic cost of walking},'' \emph{Conf Proc IEEE Eng Med Biol Soc},
  vol.~11, no.~1, pp. 3065--3068, 2014.

\bibitem{Zhang2017}
J.~Zhang, P.~Fiers, K.~A. Witte, R.~W. Jackson, K.~L. Poggensee, C.~G. Atkeson,
  and S.~H. Collins, ``{Human-in-the-loop optimization of exoskeleton
  assistance during walking},'' \emph{Science}, vol. 1284, no. June, pp.
  1280--1284, 2017.

\bibitem{Ding2018}
Y.~Ding, M.~Kim, S.~Kuindersma, and C.~J. Walsh, ``{Human-in-the-loop
  optimization of hip assistance with a soft exosuit during walking},''
  \emph{Science Robotics}, vol.~3, no.~15, pp. 1--9, 2018.

\bibitem{Panizzolo2019}
F.~A. Panizzolo, G.~M. Freisinger, N.~Karavas, A.~M. Eckert-Erdheim, C.~Siviy,
  A.~Long, R.~A. Zifchock, M.~E. LaFiandra, and C.~J. Walsh, ``{Metabolic cost
  adaptations during training with a soft exosuit assisting the hip joint},''
  \emph{Scientific Reports}, vol.~9, no.~1, pp. 1--10, 2019.

\bibitem{Sawicki2020}
G.~S. Sawicki, O.~N. Beck, I.~Kang, and A.~J. Young, ``{The exoskeleton
  expansion: Improving walking and running economy},'' \emph{J.
  NeuroEngineering and Rehabilitation}, vol.~17, no.~1, pp. 1--9, 2020.

\bibitem{Mooney2014}
L.~M. Mooney, E.~J. Rouse, and H.~M. Herr, ``{Autonomous exoskeleton reduces
  metabolic cost of human walking during load carriage},'' \emph{J.
  NeuroEngineering and Rehabilitation}, vol.~11, no.~1, pp. 1--5, 2014.

\bibitem{gordon2007learning}
K.~E. Gordon and D.~P. Ferris, ``Learning to walk with a robotic ankle
  exoskeleton,'' \emph{J. biomechanics}, vol.~40, no.~12, pp. 2636--2644, 2007.

\bibitem{lenzi2013powered}
T.~Lenzi, M.~C. Carrozza, and S.~K. Agrawal, ``Powered hip exoskeletons can
  reduce the user's hip and ankle muscle activations during walking,''
  \emph{IEEE Trans. Neural Systems and Rehabilitation Engineering}, vol.~21,
  no.~6, pp. 938--948, 2013.

\bibitem{Zhu2021}
H.~Zhu, C.~Nesler, N.~Divekar, V.~Peddinti, and R.~Gregg, ``Design principles
  for compact, backdrivable actuation in partial-assist powered knee
  orthoses,'' \emph{IEEE/ASME Trans. Mechatronics}, 2021.

\bibitem{Tucker2015}
M.~R. Tucker, J.~Olivier, A.~Pagel, H.~Bleuler, M.~Bouri, O.~Lambercy, J.~del
  R~Mill{\'a}n, R.~Riener, H.~Vallery, and R.~Gassert, ``Control strategies for
  active lower extremity prosthetics and orthotics: a review,'' \emph{J.
  neuroengineering and rehabilitation}, vol.~12, no.~1, 2015.

\bibitem{Holgate2009}
M.~A. Holgate, T.~G. Sugar, and A.~W. {\"B}ohler, ``{A novel control algorithm
  for wearable robotics using phase plane invariants},'' \emph{IEEE Int. Conf.
  Robotics and Automation}, pp. 3845--3850, 2009.

\bibitem{Quintero2018a}
D.~Quintero, D.~J. Villarreal, D.~J. Lambert, S.~Kapp, and R.~D. Gregg,
  ``{Continuous-Phase Control of a Powered Knee-Ankle Prosthesis: Amputee
  Experiments Across Speeds and Inclines},'' \emph{IEEE Trans. Robotics},
  vol.~34, no.~3, pp. 686--701, 2018.

\bibitem{Rezazadeh2019}
S.~Rezazadeh, D.~Quintero, N.~Divekar, E.~Reznick, L.~Gray, and R.~D. Gregg,
  ``{A Phase Variable Approach for Improved Rhythmic and Non-Rhythmic Control
  of a Powered Knee-Ankle Prosthesis},'' \emph{IEEE Access}, vol.~7, pp.
  109\,840--109\,855, 2019.

\bibitem{Kang2019}
I.~Kang, P.~Kunapuli, and A.~J. Young, ``{Real-Time Neural Network-Based Gait
  Phase Estimation Using a Robotic Hip Exoskeleton},'' \emph{IEEE Trans.
  Medical Robotics and Bionics}, vol.~2, no.~1, pp. 28--37, 2019.

\bibitem{Kang2021}
I.~Kang, D.~Molinaro, S.~Duggal, Y.~Chen, P.~Kunapuli, and A.~Young,
  ``Real-time gait phase estimation for robotic hip exoskeleton control during
  multimodal locomotion,'' \emph{IEEE Robotics and Automation Letters}, vol.~6,
  no.~2, pp. 3491--3497, 2021.

\bibitem{Seo2019}
K.~Seo, Y.~J. Park, J.~Lee, S.~Hyung, M.~Lee, J.~Kim, H.~Choi, and Y.~Shim,
  ``{RNN-based on-line continuous gait phase estimation from shank-mounted IMUs
  to control ankle exoskeletons},'' \emph{IEEE Int. Conf. Rehabilitation
  Robotics}, pp. 809--815, 2019.

\bibitem{hong2021phase}
W.~Hong, N.~A. Kumar, and P.~Hur, ``A phase-shifting based human gait phase
  estimation for powered transfemoral prostheses,'' \emph{IEEE Robotics and
  Automation Letters}, vol.~6, no.~3, pp. 5113--5120, 2021.

\bibitem{zhang2021adaptive}
B.~Zhang, M.~Zhou, W.~Xu \emph{et~al.}, ``An adaptive framework of real-time
  continuous gait phase variable estimation for lower-limb wearable robots,''
  \emph{Robotics and Autonomous Systems}, vol. 143, p. 103842, 2021.

\bibitem{shepherd2022deep}
M.~Shepherd, D.~Molinaro, G.~Sawicki, and A.~Young, ``Deep learning enables
  exoboot control to augment variable-speed walking,'' \emph{IEEE Robotics and
  Automation Letters}, 2022.

\bibitem{Thatte2019}
N.~Thatte, T.~Shah, and H.~Geyer, ``{Robust and adaptive lower limb prosthesis
  stance control via extended kalman filter-based gait phase estimation},''
  \emph{IEEE Robotics and Automation Letters}, vol.~4, no.~4, pp. 3129--3136,
  2019.

\bibitem{Huang2011}
H.~Huang, F.~Zhang, L.~J. Hargrove, Z.~Dou, D.~R. Rogers, and K.~B. Englehart,
  ``Continuous locomotion-mode identification for prosthetic legs based on
  neuromuscular-mechanical fusion,'' \emph{IEEE Trans. Biomedical Engineering},
  vol.~58, pp. 2867--2875, 2011.

\bibitem{Young2013}
A.~J. Young, A.~M. Simon, N.~P. Fey, and L.~J. Hargrove, ``{Classifying the
  intent of novel users during human locomotion using powered lower limb
  prostheses},'' \emph{Int IEEE EMBS Conf Neural Eng}, pp. 311--314, 2013.

\bibitem{Joshi2013}
C.~D. Joshi, U.~Lahiri, and N.~V. Thakor, ``{Classification of gait phases from
  lower limb EMG: Application to exoskeleton orthosis},'' \emph{IEEE
  Point-of-Care Healthcare Technologies}, pp. 228--231, 2013.

\bibitem{Young2016}
A.~J. Young and L.~J. Hargrove, ``A classification method for user-independent
  intent recognition for transfemoral amputees using powered lower limb
  prostheses,'' \emph{IEEE Trans. Neural Syst. Rehabilitation Eng.}, vol.~24,
  no.~2, pp. 217--225, 2016.

\bibitem{Liu2017}
M.~Liu, F.~Zhang, and H.~H. Huang, ``An adaptive classification strategy for
  reliable locomotion mode recognition,'' \emph{Sensors}, vol.~17, no.~9, 2017.

\bibitem{Hu2019}
B.~Hu, A.~M. Simon, and L.~Hargrove, ``Deep generative models with data
  augmentation to learn robust representations of movement intention for
  powered leg prostheses,'' \emph{IEEE Trans. Medical Robotics and Bionics},
  vol.~1, no.~4, pp. 267--278, 2019.

\bibitem{Embry2018}
K.~R. Embry, D.~J. Villarreal, R.~L. Macaluso, and R.~D. Gregg, ``{Modeling the
  Kinematics of Human Locomotion over Continuously Varying Speeds and
  Inclines},'' \emph{IEEE Trans. Neural Syst. Rehabilitation Eng.}, vol.~26,
  no.~12, pp. 2342--2350, 2018.

\bibitem{Embry2020}
K.~Embry and R.~D. Gregg, ``Analysis of continuously varying kinematics for
  powered prosthetic leg control,'' \emph{IEEE Transactions on Neural Systems
  and Rehabilitation Engineering}, vol.~29, pp. 262--272, 2020.

\bibitem{Camargo2021}
J.~Camargo, W.~Flanagan, N.~Csomay-Shanklin, B.~Kanwar, and A.~Young, ``{A
  Machine Learning Strategy for Locomotion Classification and Parameter
  Estimation Using Fusion of Wearable Sensors},'' \emph{IEEE Trans. Biomedical
  Engineering}, vol.~68, no.~5, pp. 1569--1578, 2021.

\bibitem{bernsteinwikipedia}
{Wikipedia contributors}, ``Bernstein polynomial --- {Wikipedia}{,} the free
  encyclopedia,''
  \url{https://en.wikipedia.org/w/index.php?title=Bernstein_polynomial&oldid=1033750467},
  2021, [Online; accessed 31-August-2021].

\bibitem{medrano2022analysis}
R.~L. Medrano, G.~C. Thomas, E.~J. Rouse, and R.~D. Gregg, ``Analysis of the
  bayesian gait-state estimation problem for lower-limb wearable robot sensor
  configurations,'' \emph{IEEE Robotics and Automation Letters}, vol.~7, no.~3,
  pp. 7463--7470, 2022.

\bibitem{VillarrealPoonawalaGregg2017TNSRE}
D.~J. Villarreal, H.~A. Poonawala, and R.~D. Gregg, ``A robust parameterization
  of human gait patterns across phase-shifting perturbations,'' \emph{IEEE
  Trans. Neural Syst. Rehabilitation Eng.}, vol.~25, no.~3, pp. 265--278, 2017.

\bibitem{Malcolm2013}
P.~Malcolm, W.~Derave, S.~Galle, and D.~{De Clercq}, ``{A Simple Exoskeleton
  That Assists Plantarflexion Can Reduce the Metabolic Cost of Human
  Walking},'' \emph{PLoS ONE}, vol.~8, no.~2, pp. 1--7, 2013.

\bibitem{MedranoThomasGreggRouse2022CodeOcean}
\BIBentryALTinterwordspacing
R.~L. Medrano, G.~C. Thomas, E.~J. Rouse, and R.~D. Gregg, ``Real-time phase
  and task estimation for controlling a powered ankle exoskeleton on extremely
  uneven terrain [{Source Code}],'' 2022. [Online]. Available:
  \url{https://doi.org/10.24433/CO.9619225.v1}
\BIBentrySTDinterwordspacing

\bibitem{julier2004unscented}
S.~J. Julier and J.~K. Uhlmann, ``Unscented filtering and nonlinear
  estimation,'' \emph{Proc. IEEE}, vol.~92, no.~3, pp. 401--422, 2004.

\bibitem{da2017supervised}
X.~Da, R.~Hartley, and J.~W. Grizzle, ``Supervised learning for stabilizing
  underactuated bipedal robot locomotion, with outdoor experiments on the wave
  field,'' in \emph{IEEE Int. Conf. Robotics and Automation}, 2017, pp.
  3476--3483.

\bibitem{peng2022actuation}
X.~Peng, Y.~Acosta-Sojo, M.~I. Wu, and L.~Stirling, ``Actuation timing
  perception of a powered ankle exoskeleton and its associated ankle angle
  changes during walking,'' \emph{IEEE Trans. Neural Systems and Rehabilitation
  Engineering}, 2022.

\bibitem{qian2022predictive}
Y.~Qian, Y.~Wang, C.~Chen, J.~Xiong, Y.~Leng, H.~Yu, and C.~Fu, ``Predictive
  locomotion mode recognition and accurate gait phase estimation for hip
  exoskeleton on various terrains,'' \emph{IEEE Robotics and Automation
  Letters}, vol.~7, no.~3, pp. 6439--6446, 2022.

\bibitem{Reznick2020}
E.~Reznick, K.~Embry, and R.~D. Gregg, ``{Predicting Individualized Joint
  Kinematics over a Continuous Range of Slopes and Speeds},'' \emph{Proc IEEE
  RAS EMBS Int Conf Biomed Robot Biomechatron}, pp. 666--672, 2020.

\bibitem{Reznick2020a}
\BIBentryALTinterwordspacing
E.~Reznick, K.~Embry, R.~Neuman, E.~Bolivar-Nieto, N.~P. Fey, and R.~D. Gregg,
  ``Lower-limb kinematics and kinetics during continuously varying human
  locomotion,'' \emph{Scientific Data}, vol.~8, no.~1, pp. 1--12, 2021.
  [Online]. Available: \url{https://doi.org/10.1038/s41597-021-01057-9}
\BIBentrySTDinterwordspacing

\end{thebibliography}
}

\end{document}